\documentclass{article}

    \PassOptionsToPackage{numbers, compress, sort&compress}{natbib}

    \usepackage[final]{tackling_climate_workshop_style}

     \usepackage{tackling_climate_workshop_style}
\usepackage[utf8]{inputenc} 
\usepackage[T1]{fontenc}    
\usepackage{hyperref}       
\usepackage{url}            
\usepackage{booktabs}       
\usepackage{amsfonts}       
\usepackage{nicefrac}       
\usepackage{microtype}      
\usepackage{graphicx}       
\usepackage{caption}
\usepackage{subcaption}
\usepackage{multirow}

\title{A machine learning pipeline \\ for automated insect monitoring}

\author{%
  Aditya Jain\thanks{Equal contribution.} \thanks{Correspondence to: \texttt{moth-ai@mila.quebec}} \\
  Mila - Quebec AI Institute\\
  \And
  Fagner Cunha\footnotemark[1]  \\
  Federal University of Amazonas \\
  \And
  Michael Bunsen\footnotemark[1] \\
  Mila - Quebec AI Institute \\
  \And
  Léonard Pasi\thanks{Work done while at Mila - Quebec AI Institute.} \\
  EPFL \\
  \And
  Anna Viklund\footnotemark[3] \\
  Daresay \\
  \And
  Maxim Larrivée \\
  Montreal Insectarium \\
  \And
  David Rolnick \\
  McGill University\\
  Mila -- Quebec AI Institute \\
}

\begin{document}

\maketitle
\begin{abstract}
Climate change and other anthropogenic factors have led to a catastrophic decline in insects, endangering both biodiversity and the ecosystem services on which human society depends. Data on insect abundance, however, remains woefully inadequate. Camera traps, conventionally used for monitoring terrestrial vertebrates, are now being modified for insects, especially moths. We describe a complete, open-source machine learning-based software pipeline for automated monitoring of moths via camera traps, including object detection, moth/non-moth classification, fine-grained identification of moth species, and tracking individuals. We believe that our tools, which are already in use across three continents, represent the future of massively scalable data collection in entomology.
\end{abstract}
\section{Introduction}
The Earth is undergoing a sixth mass extinction event, where an \textit{eighth of all species} may become extinct by 2100 \cite{ceballos2015accelerated, ceballos2017biological, ceballos2020vertebrates}.  Insects account for about half of all living species on earth and  40\% of the animal biomass \cite{stork2018many}, but both the diversity and abundance of insects are undergoing a precipitous decline \cite{wagner2020insect} as a result of several factors, in which climate change figures prominently. The ``insect apocalypse'' significantly increases the risk of breakdown of ecosystem functions on which human society depends \cite{diaz2019pervasive}.  Monitoring insects is therefore a crucial component of climate change adaptation.


\begin{figure}[h]
     \centering
     \begin{subfigure}{0.22\textwidth}
         \centering
         \includegraphics[width=0.8\textwidth]{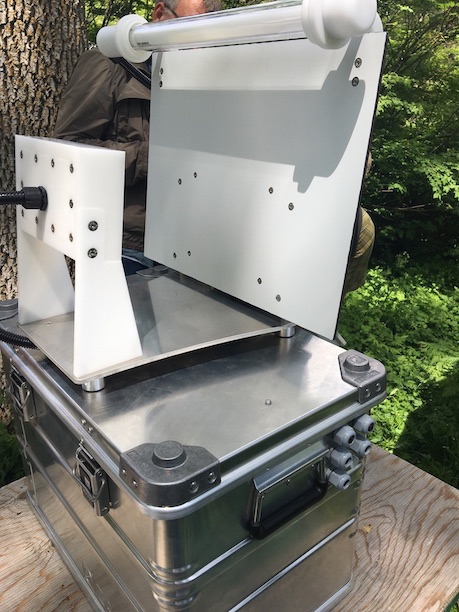}
         \caption{A moth camera trap}
         \label{fig:trap1}
     \end{subfigure}
     \begin{subfigure}{0.25\textwidth}
         \centering
         \includegraphics[width=\textwidth]{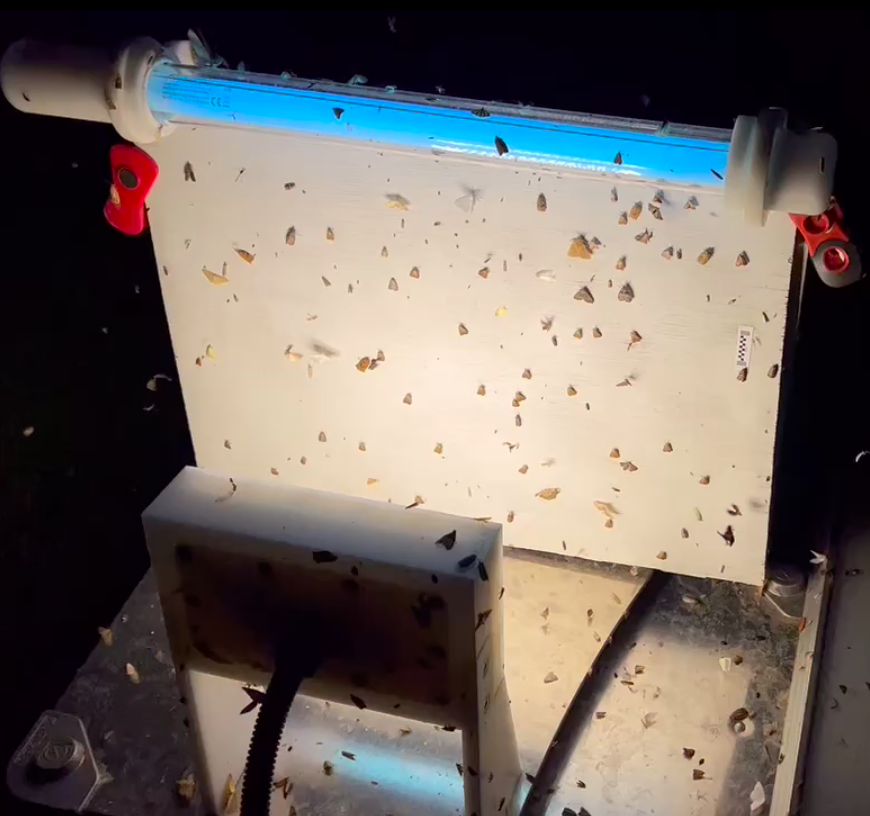}
         \caption{Trap in operation}
         \label{fig:trap2}
     \end{subfigure}     
     \begin{subfigure}{0.444\textwidth}
         \centering
         \includegraphics[width=\textwidth]{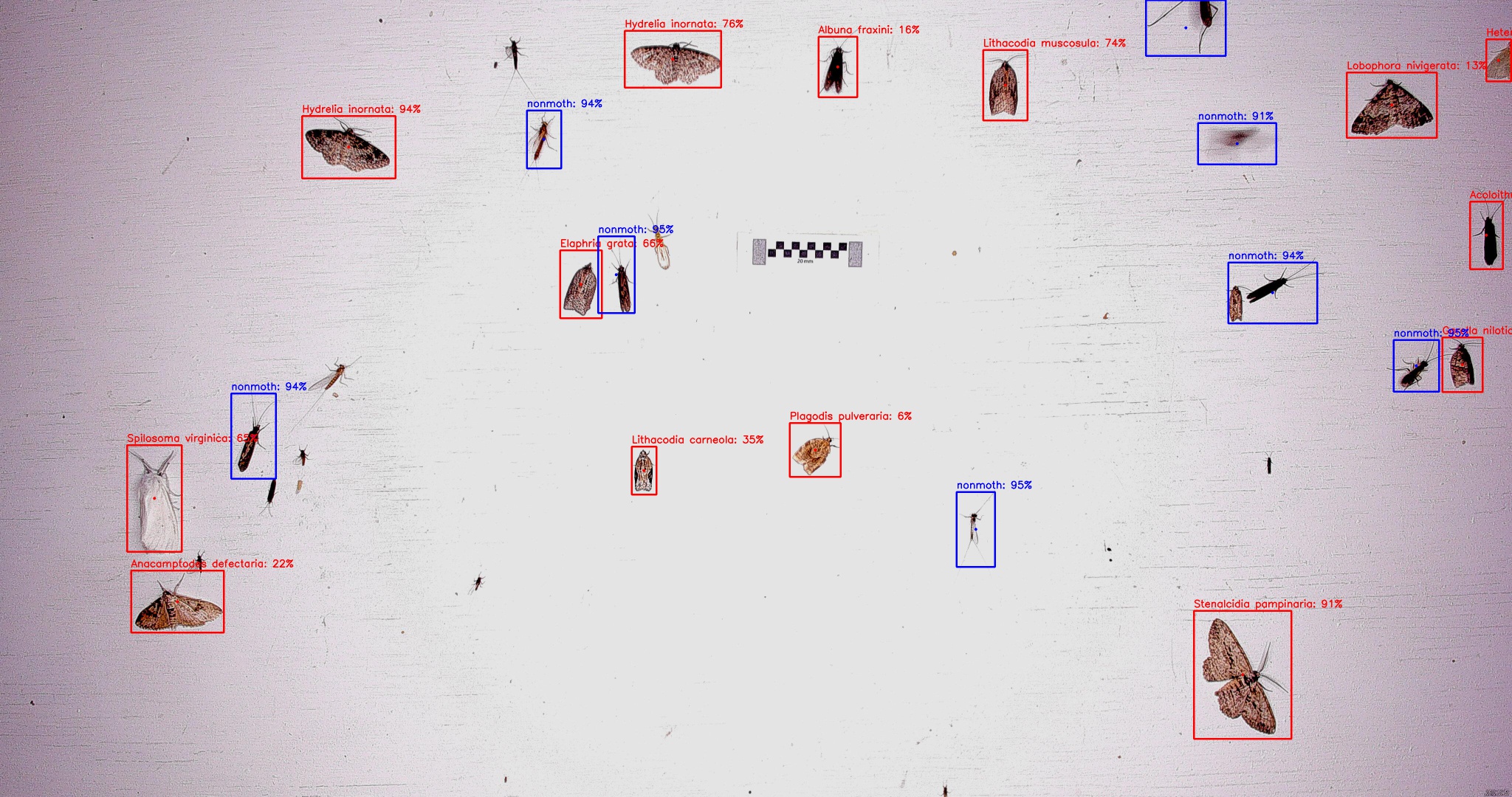}
         \caption{Machine learning predictions on a raw image}
         \label{fig:ML prediction}
     \end{subfigure}
        \caption{\autoref{fig:trap1} and \autoref{fig:trap2} depict a moth camera trap used by our partners. \autoref{fig:ML prediction} shows our insect localization and species prediction on a raw image, with red boxes showing insects classified as moths (with fine-grained species predictions), and blue boxes showing non-moths.}
        \label{fig:trap_prediction}
\end{figure}

Traditional insect collection and identification by entomologists is hard to scale, due to the massive number of insect species and a lack of experts, with certain geographies and taxonomic groups especially poorly covered. The emergence of high-resolution cameras, low-cost sensors, and processing methods based on machine learning (ML) has the potential to fundamentally change insect monitoring methods \cite{van2022emerging}. Camera traps powered by computer vision models for terrestrial vertebrates monitoring are now commonplace \cite{oliver2023camera}, and specialized camera trap hardware for insect monitoring has begun to gain momentum \cite{bjerge2022real, bjerge2023accurate, suto2022codling, geissmann2022sticky, alison2022moths}. A common group of focus for such studies has been moths\cite{bjerge2021automated, korsch2021deep, moglich2023towards}, which serve vital ecological roles and represent a fifth of all insect species. Importantly, most moths can readily be attracted with UV light and are frequently visually distinguishable up to species or genus, making them ideal targets for camera traps. As hardware for moth-monitoring has grown more common, however, there is a need for scalable data processing techniques to match the influx of data. Prior methods have been greatly limited in the species and geography covered, as well as requiring extensive manual labelling for training the algorithms.

This work describes a complete software and ML framework that transforms raw photos from insect camera traps into species-level moth data (see Fig.~\ref{fig:trap_prediction}). Our system is motivated by the following goals: 1) Model predictions should be highly accurate across moth species, 2) the presence of non-moths must be accounted for, 3) the algorithms should work across different hardware setups (camera, lighting, etc.) and geographic regions, 4) extensive manual labelling of training images should not be required, 5) the pipeline should be fast to run, 6) the system should be user-friendly for non-ML experts, and 7) the framework should easily be extended to other insect groups. Our tools are currently supporting multiple deployments in Canada, the USA, the UK, Denmark, and Panama.

\section{Machine learning pipeline}
\begin{figure}[th]
  \centering
  \includegraphics[width=\textwidth]{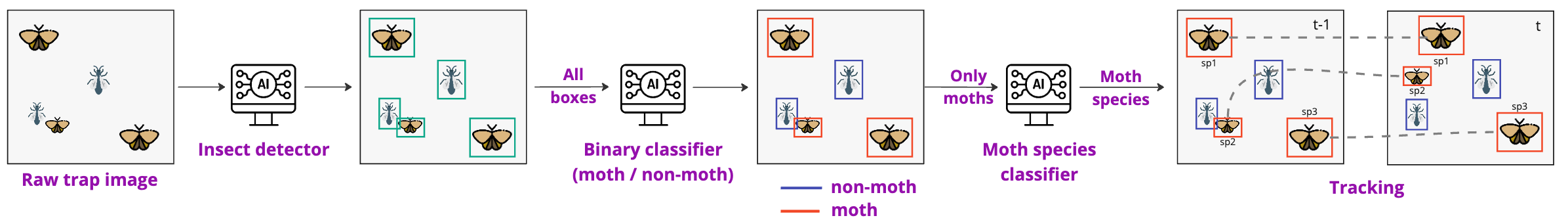}
  \caption{Machine learning workflow.
\vspace{-.2in}}
  \label{fig:ml_workflow}
\end{figure}

Our machine learning pipeline in a multi-stage process (\autoref{fig:ml_workflow}): 1) An object detector localizes all insects in the image (\S\ref{sec:insect_detector}), 2) a binary image classifier distinguishes moths from other arthropods (\S\ref{sec:binary_classifier}), 3) the classified moths pass through a fine-grained moth species classifier to predict the species (\S\ref{sec:fine_grained_classification}), and 4) a tracking algorithm tracks the moths across frames to count the number of individuals (\S\ref{sec:tracking}). Such a modular structure allows parallel development, improvement, and evaluation of each module. We now discuss each module in detail.

\subsection{Training data for image classification}
We need labelled data to train the binary and fine-grained species classifier. However, labelling data from the trap images directly is not scalable. The trap data likely would not include all rare species, labelling would be redone periodically with every change in hardware and new geographic location, and experts for species identification are scarce. We perform zero-shot transfer learning -- training on a visually different dataset from the moth trap images. Specifically, we use annotated data indexed by the Global Biodiversity Information Facility (GBIF) \cite{gbif2023}, in particular from the citizen science platform iNaturalist \cite{iNat}. Unlabeled moth trap data is provided by a network of partner ecologists.

Our pipeline for obtaining data from GBIF is as follows. Given a regional list of moth species, we compare it against the GBIF taxonomy backbone \cite{gbifbackbone}. We remove duplicate entries, merge synonym names into accepted names, and investigate doubtful, fuzzy, and unmatched names to make a processed checklist. We use the unique taxon key for each species to fetch images and metadata (such as location and publishing institution) using the Darwin Core Archive (DwC-A) \cite{dwca} file provided by GBIF. Appendix~\ref{sec:dataset_clean} discusses additional steps we apply to clean the data.

\subsection{Insect detection}\label{sec:insect_detector}
Object detection, drawing bounding boxes around objects of interest and classifying them into different categories, is now a mature problem in computer vision \cite{liu2020deep}. Our setting involves a uniform pale background, fixed camera angle and distance from the screen, which makes it possible to obtain decent performance with classical image processing techniques, such as background subtraction and blob detection \cite{bjerge2021automated}. However, the accuracy of these algorithms remains limited due to additional challenges, such as the extreme diversity of insects attracted to the detector, cases of overlap between individual insects, inconsistent lighting, and the gradual buildup of dirt on the screen. Even foundation models for object detection, such as the Segment Anything Model (SAM) \cite{kirillov2023segment}, have limitations on this relatively simple data, include failure to detect the smaller moths, cropping out legs and antennae, and low inference speed. Hence, we trained a custom model for this task.

The challenge is the lack of annotated data for the trap images to train an object detector. We use a weak annotation system leveraging SAM to generate a synthetic training dataset to minimize manual annotation. First, we use SAM to segment nearly 4k insect crops from 300 trap images from five trap deployments. Second, we review the crops to remove undesirable images, which gives us 2600 clean crops. Each crop review only takes a split second, much faster than drawing a bounding box. Third, the crops are randomly pasted on empty background images with simple augmentations (flips and rotations) to create a large simulated labelled dataset of 5k images. We train two versions of Faster R-CNN models \cite{ren2015faster} on this synthetic dataset: a slow model (ResNet-50-FPN backbone) and a fast model (MobileNetV3-Large-FPN backbone). The latter is 6 times faster than the former on a CPU while having a near-similar accuracy. Due to lack of ground truth evaluation, we visually analyze the performance of these models (App.~\ref{sec:insect_detector}).



\subsection{Moth / non-moth classification}\label{sec:binary_classifier}
Since moths are not the only arthropods that appear on the camera trap screen, we train an image classifier to differentiate between moths and other insects. Using expert knowledge of the taxa likely to appear on the screen, we fetch 350,000 images from GBIF for each moth and non-moth category. The moth group consists of adult-stage images of moth species from multiple regions worldwide. The non-moth group comprises Trichoptera, Formicidae, Ichneumonidae, Diptera, Orthoptera, Hemiptera, Pholcidae, Araneae, Opiliones, Coleoptera, and Odonata.  For this binary classification task, we train a ResNet-50 model \cite{he2016deep}, pre-trained on ImageNet-1K \cite{russakovsky2015imagenet}. The model achieves an accuracy of 96.24\% on a held-out set from GBIF data and 95.10\% on 1000 expert-annotated insect crops from camera trap images.

\subsection{Fine-grained moth species classification}\label{sec:fine_grained_classification}
The most challenging algorithm in our pipeline is for species-level identification of moths, representing a fine-grained classification task in computer vision. As there are almost 160,000 known moth species in the world~\cite{wagner2020insect}, many with very limited data, we train separate models on regional species lists. However, the task is still quite challenging, as a regional list typically contains several thousand species, and closely related species may have only subtle visual differences. Another issue is that the number of training examples per species follows a long-tail distribution, i.e., some species have many images, while most have only a few (see App. \ref{sec:species_dist}). This can bias the model towards the majority species, even if the rare species may be of particular ecological interest.

We train separate models on regional species lists using the standard ResNet-50 architecture~\cite{he2016deep}. Models are trained using AdamW optimizer~\cite{loshchilov2018decoupled}, cosine decay learning rate schedule with linear warm up~\cite{he2019bag}, and label smoothing~\cite{szegedy2016rethinking}. To minimize the performance degradation caused by the distribution shift between GBIF training data and the data from the moth traps, we apply a set of strong data augmentation operations: random crop, random horizontal flip, RandAugment~\cite{cubuk2020randaugment}, and a mixed-resolution augmentation that simulates the relatively low resolution of the cropped images from the moth traps. More details on the hyper-parameters are in App. \ref{sec:fine_grained_classification_hp}.  To mitigate the impact of majority classes on the model, we limit the number of training examples per species to 1,000. 
On a species list from the region of Quebec and Vermont, USA, our model achieves an accuracy of 86.14\% on the GBIF held-out test set, and 77.81\% on a small expert-labelled moth trap test set. We provide additional results in App. \ref{sec:fine_grained_classification_results}.

\subsection{Tracking individual moths}\label{sec:tracking}
Since the final goal is to count the number of individuals for each moth species in a night's data, tracking becomes an essential part of the system. For memory reasons, the camera trap does not continuously collect images, taking pictures at a fixed interval or when activated by insect motion. This means that movements between frames can be relatively large and "jerky".

We approach object tracking by noting that instances of the same moth in consecutive frames will likely be close to each other, similar in size, and similar in the feature space of the species classifier. Formally, we calculate the cost of the assignment between any two moth crops in two consecutive images as a weighted combination of four factors: 1) Intersection over union, 2) ratio of crop sizes, 3) distance between box centres, and 4) similarity in classification model feature space. The lower the cost, the more likely the match. We use linear sum assignment \cite{crouse2016implementing} for optimal matching of the moth crops, with unmatched individuals indicating that a moth has either appeared for the first time, or has left the image. Due to a lack of ground truth data for tracking, our evaluation for this module is currently qualitative.

\section{Pathway to impact}
\begin{figure}[hbt]
    \centering
    \begin{subfigure}{0.33\textwidth}
        \includegraphics[width=\textwidth]{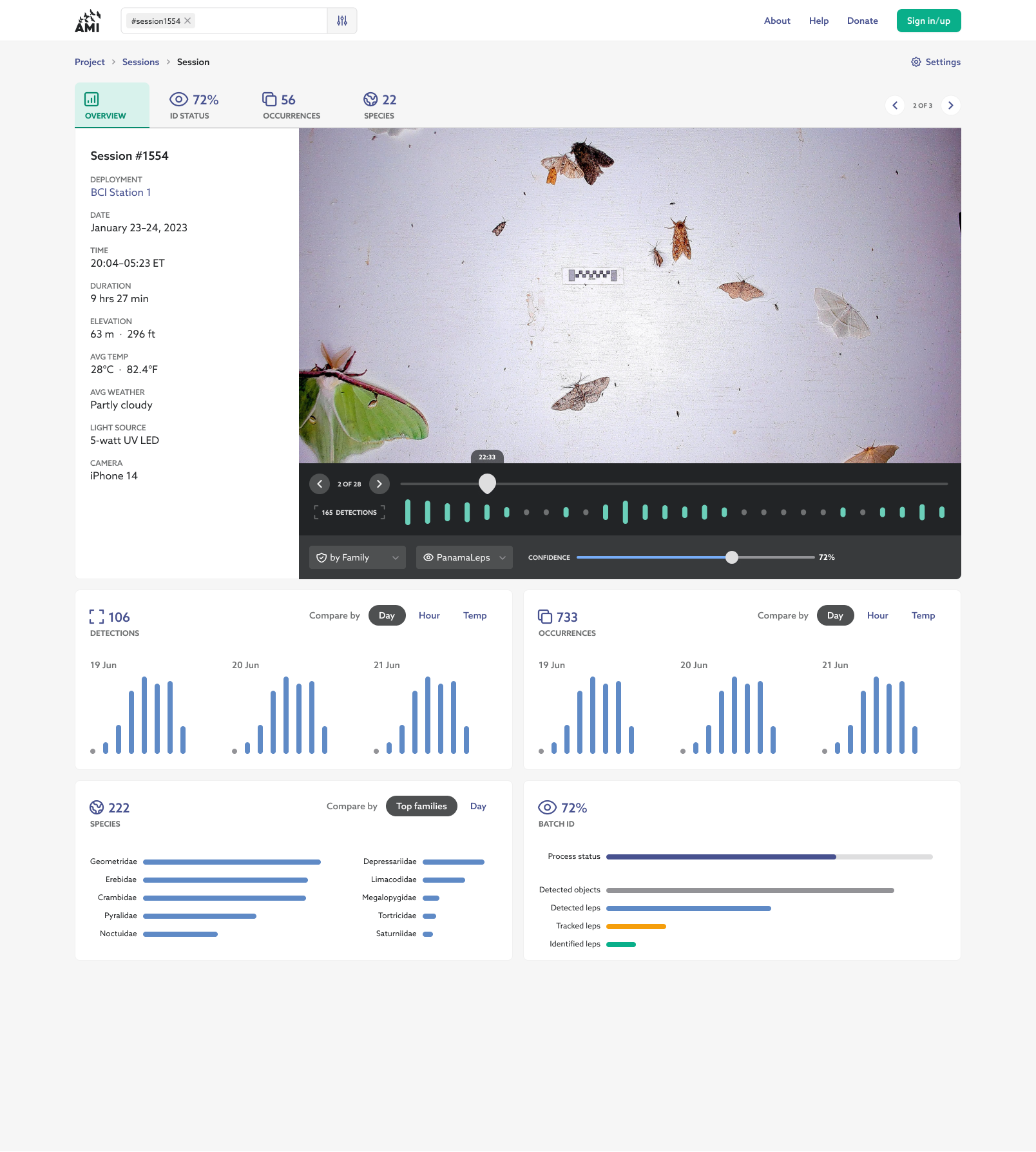}
    \end{subfigure}
    \qquad
    \begin{subfigure}{0.48\textwidth}
        \includegraphics[width=\textwidth]{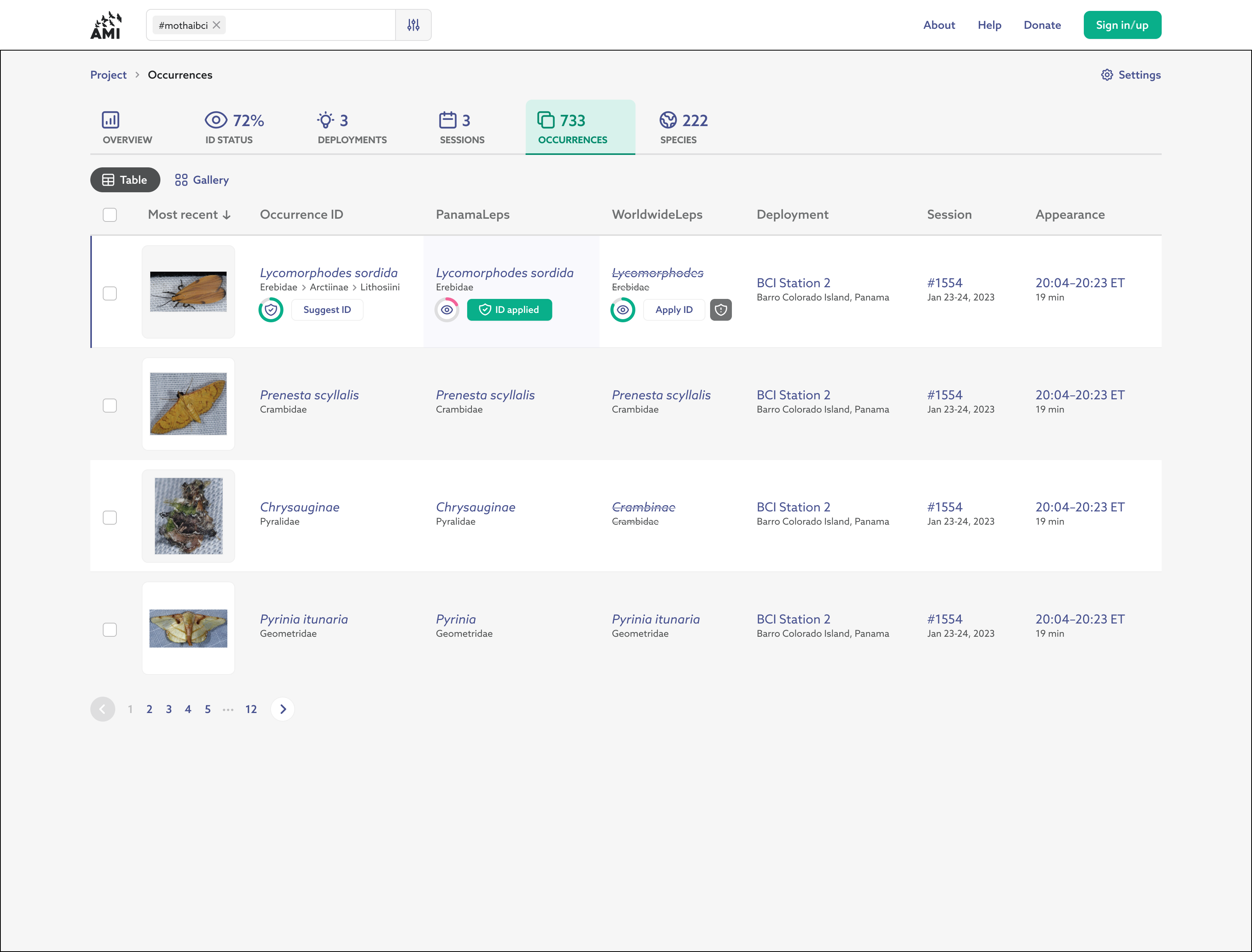}
    \end{subfigure}    
    \caption{A preview of the AMI Web Platform in development.}
    \label{fig:ami_data_companion}
\end{figure}

In order for ML tools to be usable for ecologists without extensive ML experience, we have developed a tool, the AMI Data Companion, an open-source software package (link omitted for anonymity) designed to assist researchers with these steps. (AMI stands for \textbf{A}utomated \textbf{M}onitoring of \textbf{I}nsects.) Data from field deployments of camera systems can be reviewed before processing takes place, for example, to confirm that the devices ran as scheduled. Images can be processed on demand or placed into a fault-tolerant queue for long-running operations. The choice of model is configurable for each operation (e.g.~object detection, tracking, species classification). The software provides a graphical interface that can be run on all major desktop operating systems, as well as a command-line interface that can be run on multiple server nodes in parallel. 
We are also developing an online version of the application (in progress). The AMI Web Platform (\autoref{fig:ami_data_companion}) is designed to address challenges in taxonomic alignment, access to compute infrastructure, the curation of training data, and the need to collaborate with experts from around the world. 

These tools have enabled our work to used across a range of partner ecology organizations, and we are actively scaling up deployments worldwide. It is our hope that the data our methods provide to entomologists will help better inform land use decisions and policy making for climate change adaptation and conservation.





\section*{Acknowledgement}
We sincerely thank our partners and their team without whom this project could not have been conceived and deployed. Here is a non-exhaustive list of key people: Kent McFarland\footnote[1]{Vermont Center for Ecostudies}, David Roy\footnote[2]{UK Centre for Ecology \& Hydrology\label{ceh}}, Tom August\footref{ceh}, Alba Gomez Segura\footref{ceh}, Marc Bélisle\footnote[3]{Université de Sherbrooke}, Toke Thomas Høye\footnote[4]{Aarhus University\label{aarhus}} and Kim Bjerge\footref{aarhus}. Thanks to Michelle Lin\footnote[5]{Mila - Quebec AI Institute\label{mila}} and Yuyan Chen\footref{mila} for their feedback on the manuscript.

\bibliography{references}

\appendix
\section{Appendix}
\subsection{Examples of images removed during dataset cleaning}\label{sec:dataset_clean}

\begin{figure}[htb]
    \centering
    \begin{subfigure}{0.39\textwidth}
        \includegraphics[width=\textwidth]{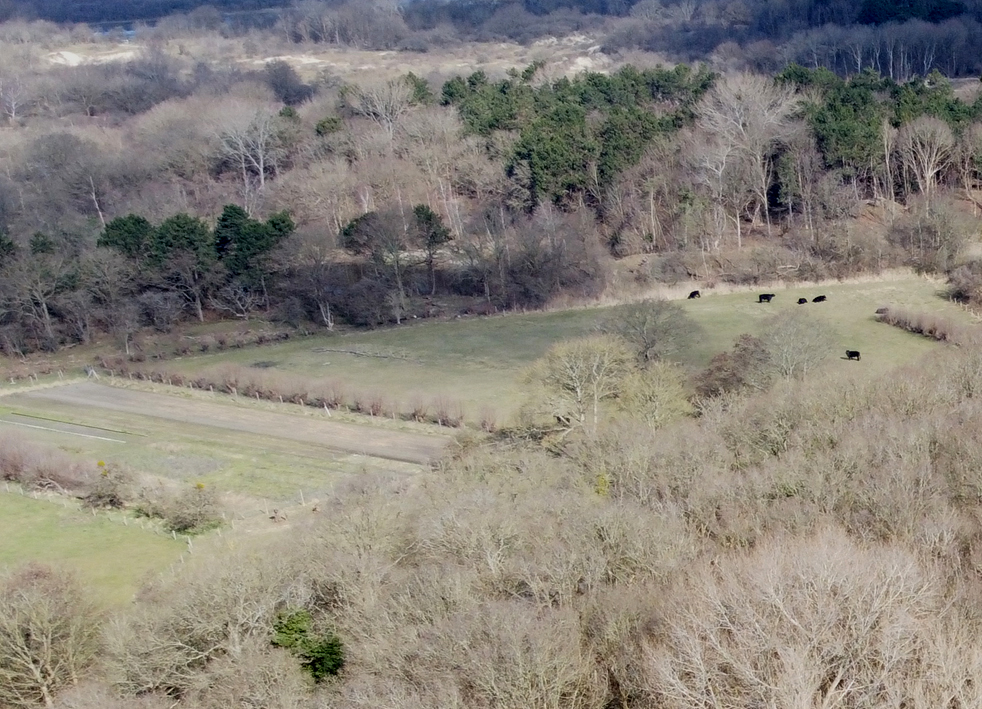}
        \caption{Habitat only}
    \end{subfigure}
    \begin{subfigure}{0.39\textwidth}
        \includegraphics[width=\textwidth]{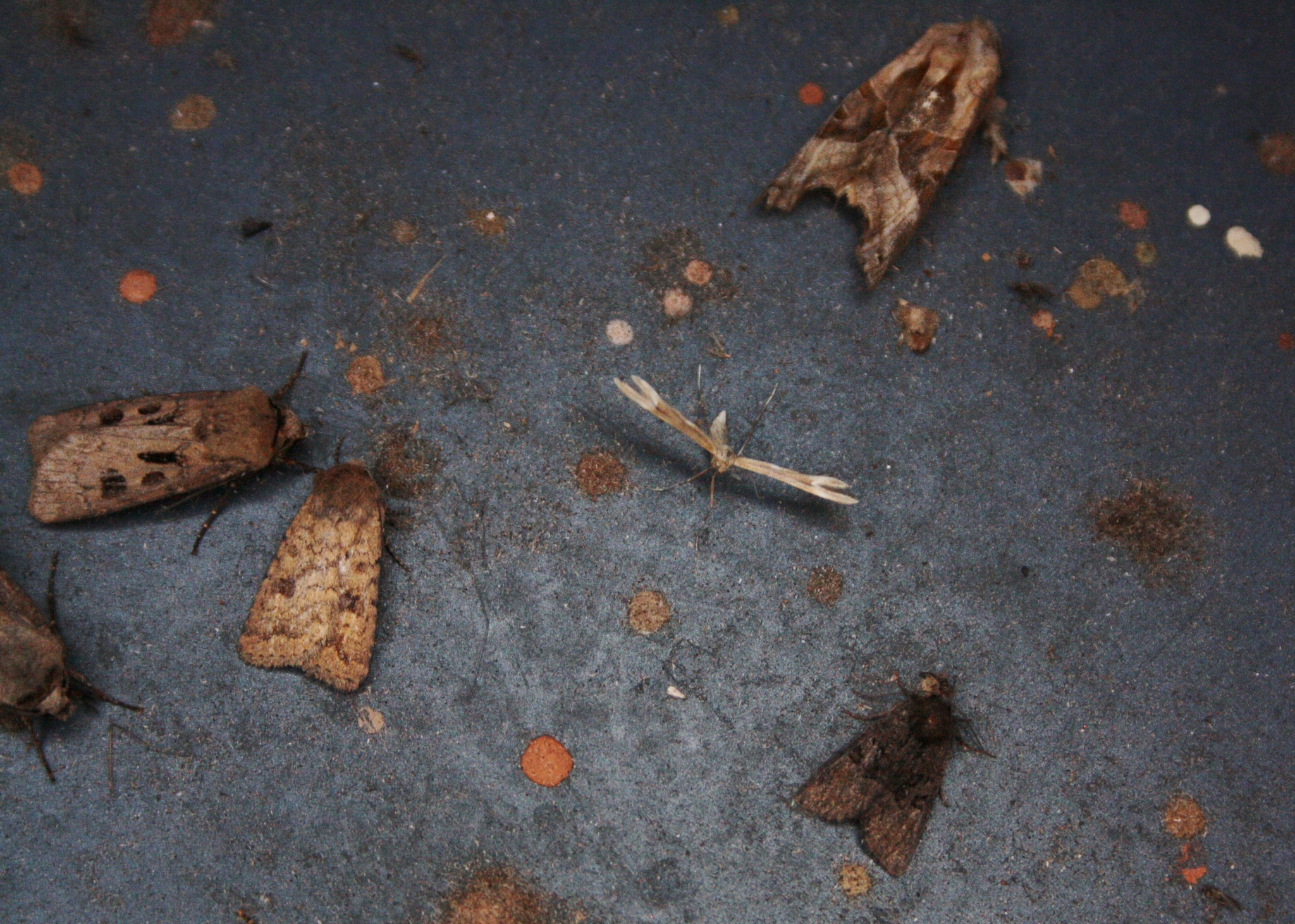}
        \caption{Duplicate}
    \end{subfigure}
    \begin{subfigure}{0.19\textwidth}
        \includegraphics[width=\textwidth]{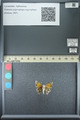}
        \caption{Thumbnail}
    \end{subfigure}
    \begin{subfigure}{0.39\textwidth}
        \includegraphics[width=\textwidth]{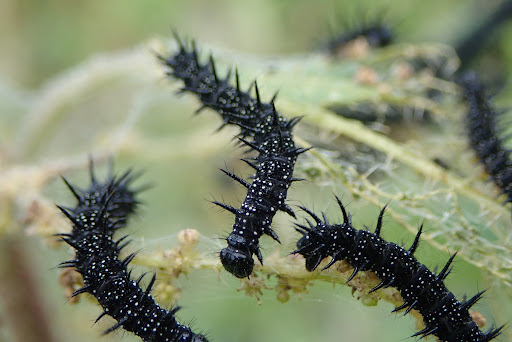}
        \caption{Non-adult}
    \end{subfigure}
    \begin{subfigure}{0.58\textwidth}
        \includegraphics[width=\textwidth]{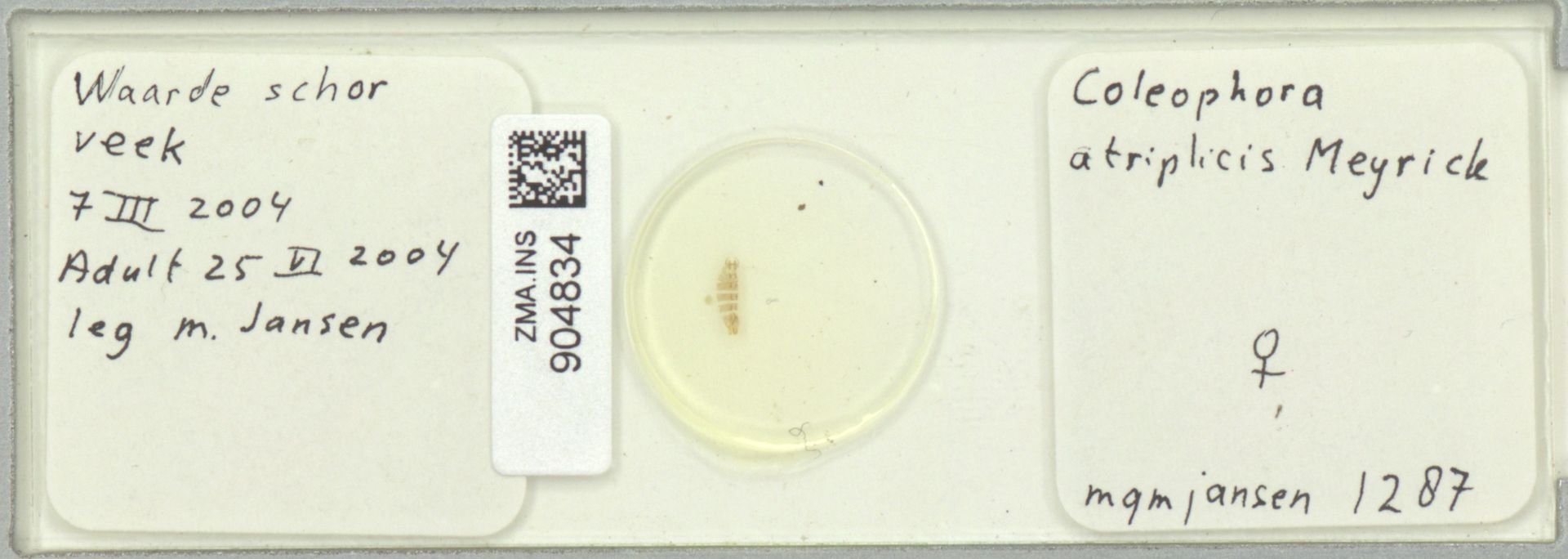}
         \caption{Description only}
    \end{subfigure}
    \caption{Examples of images that are removed during the dataset cleaning procedure. Some images are used as placeholders and do not contain any animals. (a) is an extreme case used by hundreds of thousands of occurrences. In some cases, the same picture has more than one species, and each individual is counted as a single occurrence, with the same image being referenced by all of them (b). Some occurrences have more than one image, and some of them are small images (thumbnails) (c). The scope of our models is only the adult individuals; non-adult pictures (d) should be removed. Finally, some pictures do not have any specimens, only descriptions (e).}
    \label{fig:removed_images}
\end{figure}

After fetching images and metadata, we clean the data by removing the following images: duplicates, thumbnails, non-adult images (based on metadata or using a life-stage classifier when metadata is not available), and images from datasets that we have identified as containing primarily descriptive information rather than specimen pictures. The \autoref{fig:removed_images} shows some examples of images we removed during our dataset cleaning procedure.

\subsection{Insect detector analysis}\label{sec:insect_detector_analysis}
We initially thought blob detection should have good performance for this problem setting, given its relative simplicity. However, it fails when there is a lot of insect activity on the screen or inconsistent lightning. We cherry-pick images with a clean background and low density of insects to generate annotations using blob detection automatically. We then train a Faster R-CNN with a ResNet-50-FPN backbone (blobdata-ResNet50) on this data. In addition to being slow, this model is prone to two types of errors: missed detections (i.e. false negatives), especially on smaller moths, and double detections (i.e. bounding boxes that group multiple moths). Additionally, the model occasionally made multiple predictions on the same large moth.

\begin{figure}[h]
     \centering
     \begin{subfigure}{0.3\textwidth}
         \centering
         \includegraphics[width=\textwidth]{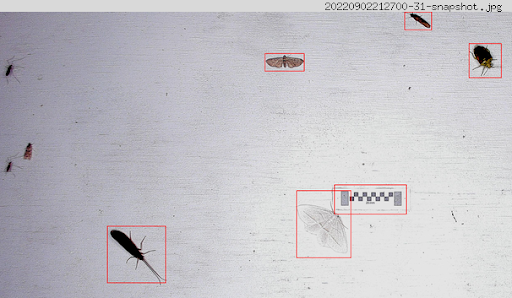}
     \end{subfigure}
     \begin{subfigure}{0.3\textwidth}
         \centering
         \includegraphics[width=\textwidth]{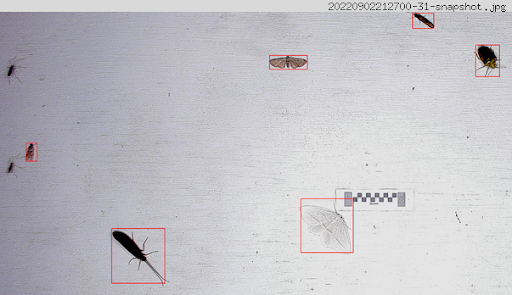}
     \end{subfigure}
     \begin{subfigure}{0.3\textwidth}
         \centering
         \includegraphics[width=\textwidth]{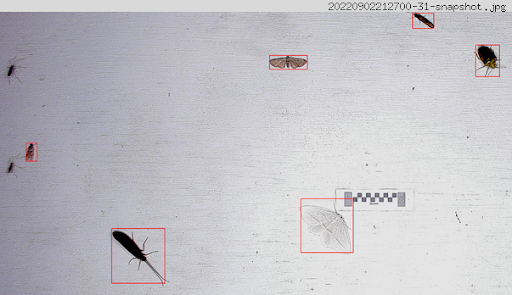}
     \end{subfigure}
     \begin{subfigure}{0.3\textwidth}
         \centering
         \includegraphics[width=\textwidth]{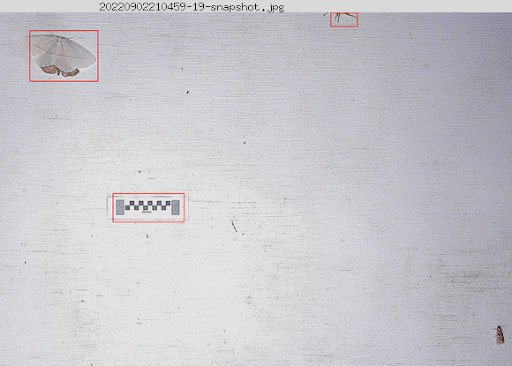}
     \end{subfigure}
     \begin{subfigure}{0.3\textwidth}
         \centering
         \includegraphics[width=\textwidth]{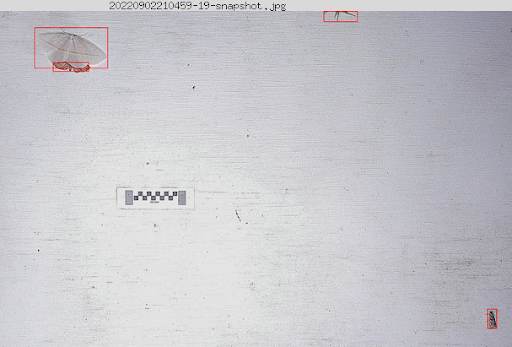}
     \end{subfigure}
     \begin{subfigure}{0.3\textwidth}
         \centering
         \includegraphics[width=\textwidth]{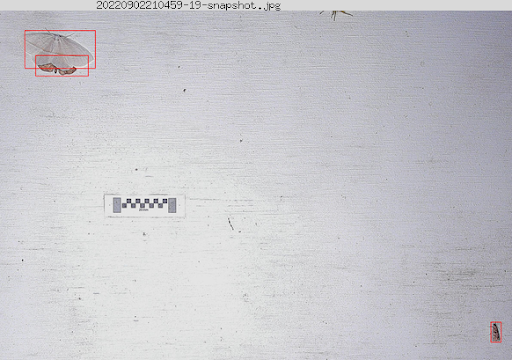}
     \end{subfigure}
     \begin{subfigure}{0.3\textwidth}
         \centering
         \includegraphics[width=\textwidth]{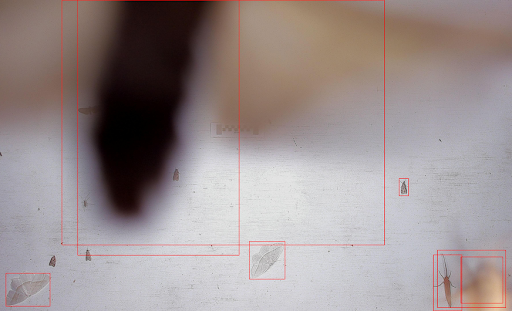}
     \end{subfigure}
     \begin{subfigure}{0.3\textwidth}
         \centering
         \includegraphics[width=\textwidth]{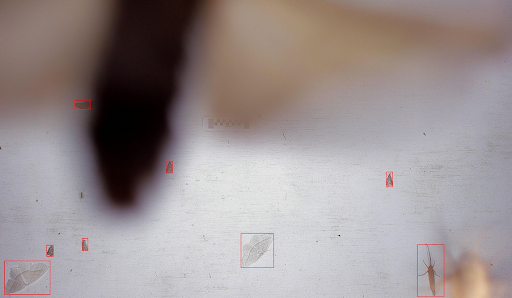}
     \end{subfigure}
     \begin{subfigure}{0.3\textwidth}
         \centering
         \includegraphics[width=\textwidth]{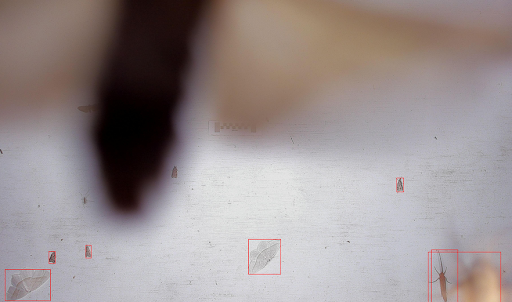}
     \end{subfigure}
    \begin{subfigure}{0.3\textwidth}
         \centering
         \includegraphics[width=\textwidth]{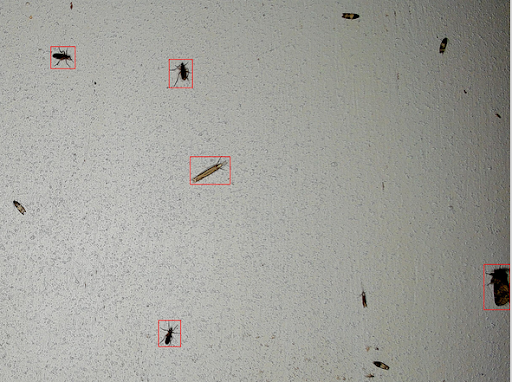}
     \end{subfigure}
     \begin{subfigure}{0.3\textwidth}
         \centering
         \includegraphics[width=\textwidth]{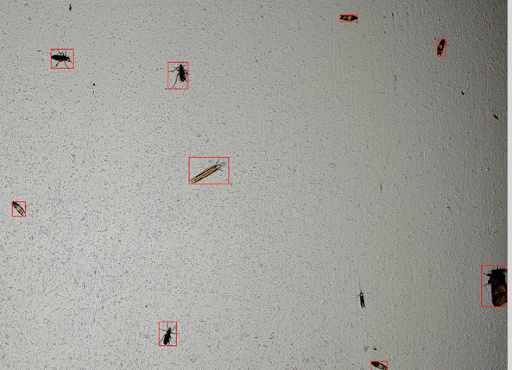}
     \end{subfigure}
     \begin{subfigure}{0.3\textwidth}
         \centering
         \includegraphics[width=\textwidth]{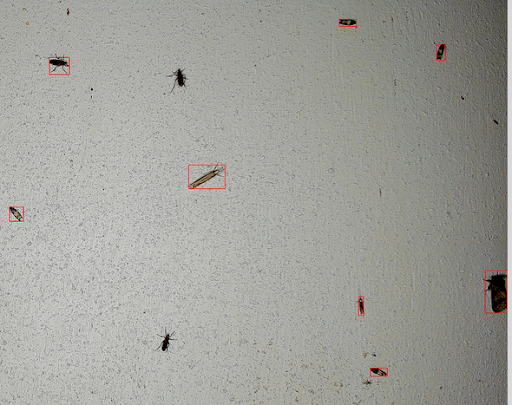}
     \end{subfigure}
     \begin{subfigure}{0.3\textwidth}
         \centering
         \includegraphics[width=\textwidth]{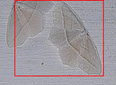}
     \end{subfigure}
     \begin{subfigure}{0.3\textwidth}
         \centering
         \includegraphics[width=\textwidth]{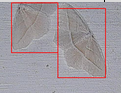}
     \end{subfigure}
     \begin{subfigure}{0.3\textwidth}
         \centering
         \includegraphics[width=\textwidth]{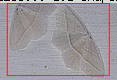}
     \end{subfigure}
     \caption{Left column: blobdata-ResNet50; middle column: syntheticdata-ResNet50; right column: syntheticdata-MobileNetV3. As seen in the first column, the model trained on a small amount of labelled data has many false positives and fails to detect insects close to each other. While the models trained on large amounts of synthetic data (last two columns) overcomes those challenges. We finally use the model with the MobileNetV3-Large-FPN backbone, as it is six times faster than its counterpart and similar in accuracy.}
     \label{fig:insect_detector_analysis}
\end{figure}

We attribute the above issues to a need for diversity in training data. Hence, as discussed in \autoref{sec:insect_detector}, we train two Faster R-CNN models on synthetic data generated using SAM: one with a heavier ResNet-50-FPN backbone (syntheticdata-ResNet50) and one with a lighter MobileNetV3-Large-FPN backbone (syntheticdata-MobileNetV3). \autoref{fig:insect_detector_analysis} shows a representative sample of the differences using visual inspection.

\subsection{Number of images per species distribution}\label{sec:species_dist}

\begin{figure}[htb]
    \centering
    \includegraphics[width=0.6\textwidth]{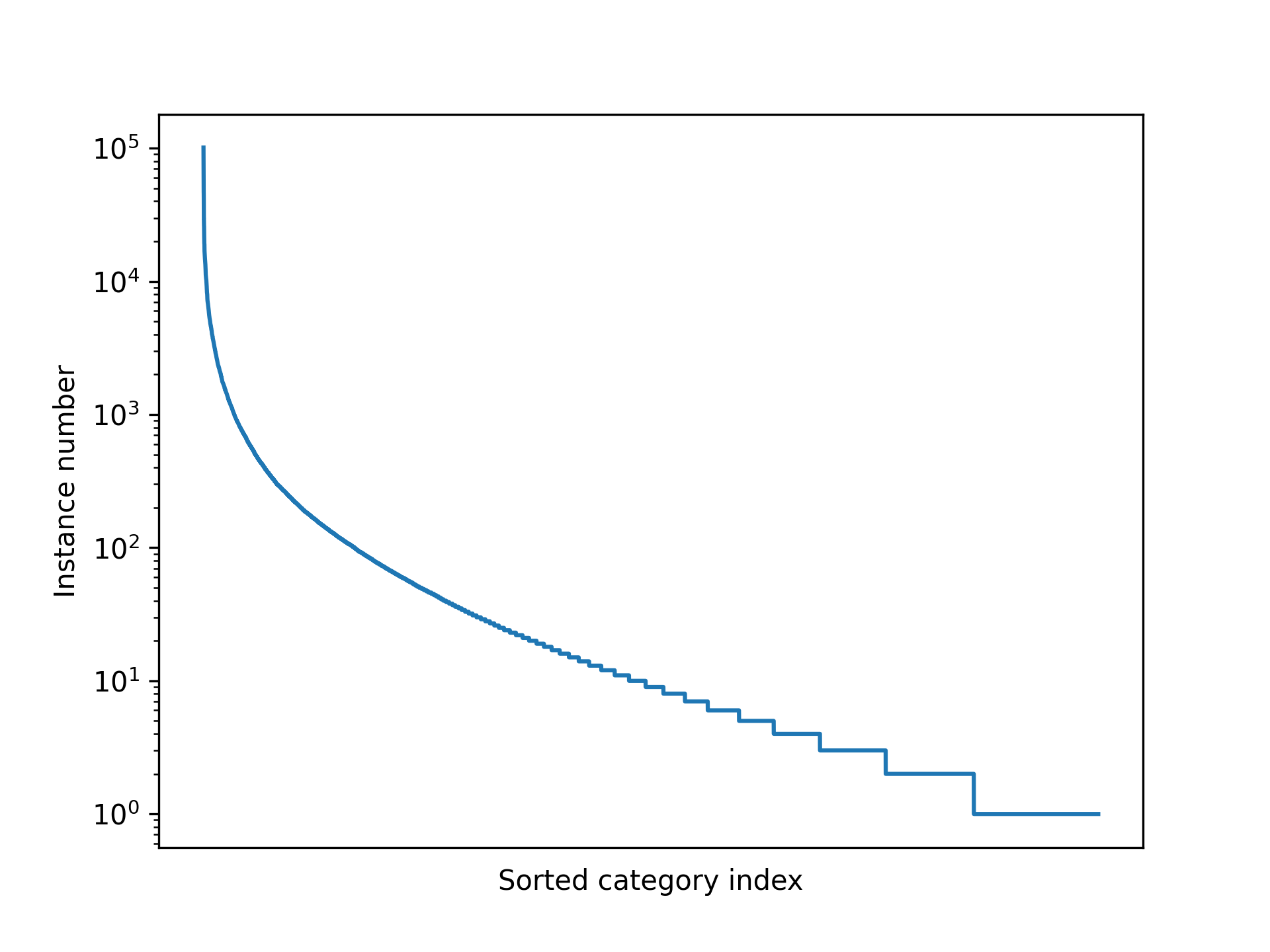}
    \caption{Number of images per species available in GBIF for the order Lepidoptera. The picture includes data from occurrences labelled as `adult' from approximately 37,000 species. Categories are sorted by the number of images.}
    \label{fig:species_dist}
\end{figure}

The number of images per species on GBIF for the Lepidoptera order follows a long-tail distribution, i.e., some species have many photos, while most have only a few, as shown in the \autoref{fig:species_dist}.

\subsection{Species classifier hyper-parameters}\label{sec:fine_grained_classification_hp}

Our classification algorithm uses the standard ResNet-50~\cite{he2016deep} architecture, which is initialized with weights pre-trained on ImageNet-1K~\cite{russakovsky2015imagenet}. We use a low input resolution of 128 x 128, which is roughly the mean size of the cropped images. We found that this resolution produces better accuracy on moth trap images. The detailed training hyper-parameters are provided in \autoref{table:classification_hp}.

\begin{table}[htb]
\begin{center}
\caption{Training hyper-parameters for fine-grained species classification task.}
\label{table:classification_hp}
\begin{tabular}{ll}
\hline
Hyper-parameters & Config \\ \hline
input resolution & 128 x 128 \\
optimizer & AdamW \\
learning rate & 0.001 \\
LR schedule & cosine \\
warmup epochs & 2 \\
training epochs & 30 \\
weight decay & 1e-5 \\
RandAug & N=2, M=9 \\
label smooth & 0.1 \\ \hline
\end{tabular}
\end{center}
\end{table}

\subsection{Classifier validation results}\label{sec:fine_grained_classification_results}

We provide validation results in \autoref{table:classifier_result} for two regional lists: Quebec-Vermont and UK-Denmark. However, it is important to note that the expert-labelled moth trap test set has only 1000 examples, with only 338 crops labelled at the species level, limiting the conclusions regarding our methods' generalisation. Obtaining more labelled data is one of our next steps.

\begin{table}[htb]
\begin{center}
\caption{Test results for Quebec-Vermont and UK-Denmark regional lists. GBIF test sets are held-out from the GBIF training images. The expert-labelled moth trap test set for Quebec-Vermont contains 1000 examples, with 338 crops labelled at the species level. We predict at a higher taxonomic level (genus and family) by summing up the confidence of predictions within each higher taxon.}
\label{table:classifier_result}
\begin{tabular}{llccc}
\hline
\multicolumn{1}{c}{\multirow{2}{*}{Dataset}} & \multicolumn{1}{c}{\multirow{2}{*}{Test set}} & \multicolumn{3}{c}{Accuracy} \\
\multicolumn{1}{c}{} & \multicolumn{1}{c}{} & Species & Genus & Family \\ \hline
\multirow{2}{*}{Quebec-Vermont} & GBIF & 86.26\% & 90.20\% & 94.77\% \\
 & Moth trap & 77.58\% & 77.00\% & 89.61\% \\ \hline
UK-Denmark & GBIF & \multicolumn{1}{l}{88.77\%} & 92.43\% & 96.21\% \\ \hline
\end{tabular}
\end{center}
\end{table}

\end{document}